\documentclass{article}

\PassOptionsToPackage{numbers,sort}{natbib}
\PassOptionsToPackage{table}{xcolor}

\usepackage{iclr2027_conference,times}

\iclrfinalcopy

\usepackage{amsmath,amssymb,amsfonts}
\usepackage{amsthm}
\usepackage{nicefrac}
\usepackage{microtype}

\usepackage{graphicx}
\usepackage{booktabs}
\usepackage{multirow}
\usepackage{array}
\usepackage{colortbl}
\usepackage{wrapfig}

\usepackage{algorithm}
\usepackage{algpseudocode}

\usepackage{enumitem}
\usepackage{xspace}
\usepackage{xcolor}
\usepackage{titlesec}

\usepackage{fontawesome5}

\newcommand{\githubrepo}[1]{%
  \href{#1}{\faGithub\ \texttt{Code}}%
}

\usepackage{url}
\usepackage{hyperref}

\usepackage{amsmath,amsfonts,bm}

\def\figref#1{figure~\ref{#1}}

\def\secref#1{section~\ref{#1}}

\def\eqref#1{equation~\ref{#1}}

\def\1{\bm{1}}

\DeclareMathAlphabet{\mathsfit}{\encodingdefault}{\sfdefault}{m}{sl}
\SetMathAlphabet{\mathsfit}{bold}{\encodingdefault}{\sfdefault}{bx}{n}

\setcitestyle{numbers,square,comma}

\titlespacing*{\section}
  {0pt}
  {1.2ex}
  {0.6ex}

\definecolor{mathbg}{RGB}{252,235,235}
\definecolor{genbg}{RGB}{235,242,252}
\definecolor{medbg}{RGB}{237,248,237}
\definecolor{avgbg}{RGB}{245,245,245}

\newcolumntype{B}{>{\columncolor{genbg}}c}
\newcolumntype{G}{>{\columncolor{medbg}}c}
\newcolumntype{R}{>{\columncolor{mathbg}}c}

\newcommand{\sourcebox}[1]{%
  {\setlength{\fboxsep}{1pt}%
   \colorbox{genbg}{#1}}%
}

\newcommand{\indomainbox}[1]{%
  {\setlength{\fboxsep}{1pt}%
   \colorbox{medbg}{#1}}%
}

\newcommand{\outdomainbox}[1]{%
  {\setlength{\fboxsep}{1pt}%
   \colorbox{mathbg}{#1}}%
}

\newcommand{\ours}{\textsc{ESRL}\xspace}

\newcommand{\tabref}[1]{Table~\ref{#1}}

\newcommand{\appref}[1]{Appendix~\ref{#1}}

\renewcommand{\figref}[1]{Fig.~\ref{#1}}
\renewcommand{\secref}[1]{\S\ref{#1}}

\hypersetup{
    colorlinks=true,
    linkcolor=red,
    citecolor=cyan,
    filecolor=magenta,      
    urlcolor=cyan,
}

\title{
Continual Test-Time Learning with Evolution Strategies
}

\author{
Guangya Hao\textsuperscript{1},\quad
Yiting Duan\textsuperscript{3},\quad
Longyan Tan\textsuperscript{1},\quad
Yunbo Long\textsuperscript{1},\quad
Hao Wu\textsuperscript{3},
\\[0.3em]
\bfseries
~Ge Zheng\textsuperscript{1},\quad
Alexandra Brintrup\textsuperscript{1,4},\quad
Zhuokai Zhao\textsuperscript{2}
\\[0.6em]
\normalfont
\textsuperscript{1}University of Cambridge,
\quad
\textsuperscript{2}University of Chicago
\quad
\textsuperscript{3}Western Sydney University
\quad\\
\textsuperscript{4}The Alan Turing Institute,
\\[0.3em]
{\normalfont\small
~Correspondence:
\texttt{gh540@cam.ac.uk} and
\texttt{zhuokai@uchicago.edu}
}
}

\begin{document}

\maketitle

\begin{center}
    \vspace{-2em}
    \githubrepo{https://github.com/gh540-svg/ESRL}
\end{center}

\fancyhead[L]{Preprint}


\begin{abstract}

Large language models increasingly operate in open-ended environments, where they are expected to continually adapt to evolving tasks and distributions, often without access to external supervision. 
Existing continual learning approaches rely on external supervision, such as labeled data, demonstrations, or reward signals, which can be costly or unavailable in practice. 
Test-time adaptation removes this dependence by learning from self-generated supervision, but existing methods derive that supervision from the single policy being updated, so once the policy drifts, its own learning signal drifts with it, leading to policy collapse, degraded reasoning, and catastrophic forgetting.
In this paper, we propose \textbf{Evolution Strategy Reinforcement Learning (\ours)}, a continual test-time learning framework that draws its self-generated supervision from a population of policies instead of the single policy being trained. 
\ours perturbs an anchor policy in parameter space to form a population of nearby policies, samples one trajectory from each, and aggregates their answers into label-free group-relative learning signals for the central student.
The anchor is held fixed within each block and refreshed from the student at block boundaries, so the population stays centered on a stable yet current policy and continues to supply diverse and reliable supervision as the student moves.
In single-task test-time adaptation across mathematical reasoning, general knowledge, and medical domains, \ours improves average accuracy by up to 31.7\% relative to the base model and 16.6\% relative to the strongest test-time adaptation baseline. 
Notably, \ours substantially reduces catastrophic forgetting and achieves the strongest final performance in continual learning.

\end{abstract}


\section{Introduction}
\label{sec:intro}

Large Language models (LLMs) are deployed as autonomous agents that increasingly interact with users, tools, and environments over extended periods of time~\citep{hao2026decentmem,zhou2026omniopd}. 
Unlike conventional offline settings, these open-ended environments continuously introduce new tasks, evolving objectives, and previously unseen problems. 
To remain effective after deployment, language agents must continually improve their reasoning abilities from ongoing experience rather than relying solely on static pre-training datasets~\citep{shenfeld2026self,hao2026spd,zhou2026sage, meng2026rethinking}.  
Continual learning provides a natural way to support such adaptation by enabling models to acquire new capabilities while retaining previously learned ones~\citep{wu2024continual,zheng2025lifelong}.
However, existing approaches primarily depend on external supervision, such as labeled data~\citep{wu2024continual}, demonstrations~\citep{shenfeld2026self}, or reward signals~\citep{wu2024continual}. 
As language agents are deployed at scale, obtaining such supervision for every newly encountered task becomes increasingly impractical~\citep{hao2026spd}.
This naturally raises a fundamental question: Can a language model continually improve its reasoning at test time, without external supervision? 
Test-time learning provides a promising step toward this goal by allowing models to adapt using only unlabeled self-generated data encountered after deployment.
Recent work has demonstrated that self-generated supervision can drive reasoning capabilities without requiring external labels~\citep{zuo2025ttrl, zhou2025evolving,zhao2026echo,liu2025ettrl}. 
However, self-improvement introduces a new challenge.
When a model repeatedly learns from its own outputs, future updates become increasingly dependent on previous generations.
Over time, the resulting self-reinforcing feedback loop can amplify erroneous pseudo-labels, reduce reasoning diversity~\citep{liu2025ettrl}, and progressively collapse exploration toward a narrow set of reasoning behaviours~\citep{zhou2025evolving}.
As adaptation continues, previously acquired capabilities may be overwritten, leading to severe catastrophic forgetting~\citep{zuo2025ttrl}. 
Consequently, the central challenge is no longer simply whether a model can learn without supervision, but \emph{whether it can continually improve its reasoning while preserving the diversity necessary for sustained exploration and retaining previously acquired capabilities}. 
To address this challenge, we propose Evolution Strategy Reinforcement Learning (\ours), a continual test-time learning framework that draws its self-generated supervision from a population
of policies instead of the single policy being trained.
More specifically, \ours constructs a population of locally perturbed policies around a shared anchor policy. Then each perturbed policy independently generates a solution to the same prompt, enabling exploration across multiple nearby regions of parameter space and producing diverse reasoning trajectories. To maintain both stability and continual adaptation, \ours keeps the anchor fixed within each block and periodically refreshes it with the current student at block boundaries, enabling exploration to remain centered around an evolving yet stable reference. By maintaining policy-level diversity throughout adaptation and transferring robust consensus signals to a continuously updated student model, \ours  enables effective continual test-time learning while mitigating catastrophic forgetting. 

\begin{figure}[!t]  
    \centering
    \includegraphics[width=1\linewidth]{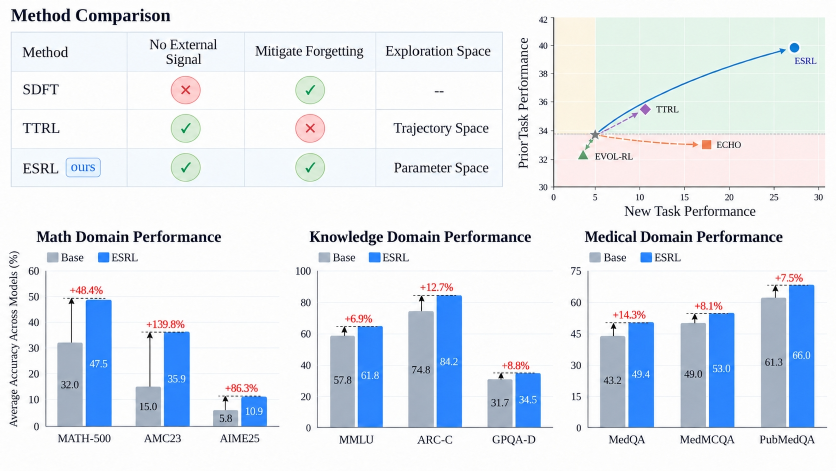}  
    \vspace{-0.4in}
    \caption{\ours comparisons, continual learning performance, and overall performance summary. Top left: Comparison with existing methods for continual test-time learning across three key dimensions. Top right: Continual learning performance after training on MATH-500. Bottom: \ours consistently improves the base models across domain-specific benchmarks and in-domain transfer tasks. All results are reported as accuracy (\%).}
    \label{fig:introduction}
    \vspace{-0.3in}
\end{figure}

Extensive experiments across four backbone models and nine benchmarks show that ESRL consistently improves test-time adaptation across mathematical reasoning, general knowledge, and medical reasoning tasks. 
In single-task test-time adaptation, \ours improves average accuracy by up to 31.7\% relative to the base model and 16.6\% relative to the strongest test-time adaptation baseline. 
In continual learning, ESRL better retains previously acquired capabilities and achieves the best final performance across all three domains, outperforming the strongest test-time adaptation baseline by up to 90\% at the end of training.

To summarize, our contributions are outlined below:
\begin{itemize}[topsep=0pt,leftmargin=*,noitemsep]
    \item We propose \ours, a continual test-time learning framework that performs policy-level exploration without external supervision. \ours constructs a population of locally perturbed policies and learns from their collective self-generated feedback.
    \item \ours introduces three key designs for stable and effective adaptation: \textit{parameter-space exploration} to broaden the reachable reasoning space, \textit{a group-relative learning objective} to aggregate signals across the policy population, and \textit{periodic anchor block updates} to consolidate newly acquired capabilities while preserving a stable reference policy.
    \item Experiments show that \ours achieves sustained reasoning gains under label-free adaptation and consistently maintains stronger performance throughout long-horizon continual learning.
\end{itemize}

\section{Related Work}
\label{sec:related_work}

\subsection{Test Time Learning}
Test-time learning adapts a model directly using unlabeled data encountered after deployment. Recent work has extended this idea to large language models through self-generated supervision. TTRL~\citep{zuo2025ttrl} uses majority voting over sampled responses to construct label-free rewards for test-time reinforcement learning. Subsequent work has mainly focused on improving the reliability and diversity
of these self-generated rollouts. ETTRL~\citep{liu2025ettrl} introduces
entropy-guided tree rollouts to improve exploration and uses advantage reshaping to reduce early overconfidence. EVOL-RL~\citep{zhou2025evolving}introduces novelty-aware rewards to counter the loss of reasoning diversity caused by majority-only optimization. ECHO~\citep{zhao2026echo} further combines entropy and confidence during rollout generation and optimization to reduce rollout collapse and noisy pseudo-label reinforcement. More recent methods further target the fragility of majority-based supervision. Hi-TTRL~\citep{xu2026hittrl} regulates rollout consensus to avoid unreliable or vanishing learning signals, and TTPO~\citep{wang2026ttpo} applies different objectives to trajectories that agree or disagree with the pseudo-label.

Despite these advances, existing methods still construct self-supervision primarily from trajectories generated by the current policy. Once the policy drifts toward a narrow or erroneous region, its self-generated signals may reinforce the same behaviour. Moreover, these methods mainly focus on the current test distribution and do not explicitly preserve prior capabilities under sequential adaptation. \ours addresses both limitations through parameter-space population exploration for more diverse and stable continual test-time learning.

\subsection{Continual Learning}

Continual learning aims to acquire new capabilities while preserving previously learned ones~\citep{wu2024continual,zhao2024direct,zheng2025lifelong}. 
Existing approaches mainly mitigate catastrophic forgetting through replay, regularization, or parameter isolation. Continual fine-tuning can cause LLMs to forget previously learned knowledge and reasoning skills~\citep{luo2025empirical}. Replay-based methods reduce forgetting by training again on data from earlier tasks, while O-LoRA and LLaMA-Pro limit interference by separating or expanding the trainable parameters~\citep{wang2023orthogonal,wu2024llamapro}. Transfer-Selective Replay~\citep{meng2026rethinking} further selects which past examples to replay based on their compatibility with the current task. Recent studies suggest that reinforcement-based adaptation can preserve previously learned capabilities better than supervised fine-tuning during sequential post-training~\citep{lai2025reinforcement}. This advantage has been linked to the use of on-policy data and the smaller distributional shift it induces~\citep{shenfeld2026rlsrazor,chen2026retaining}. Building on this idea, SDFT~\citep{shenfeld2026self} generates on-policy training signals from expert demonstrations through self-distillation, substantially reducing forgetting across sequential tasks. However, it still assumes access to demonstrations for each new task.

Despite progress in capability retention, existing continual learning methods generally rely on stored data, demonstrations, or other external supervision when new tasks arrive. They therefore do not directly address continual adaptation from unlabeled test-time data, where the model must generate its own learning signals while avoiding catastrophic forgetting. \ours bridges this gap by enabling label-free continual test-time adaptation while maintaining diverse policy-level supervision to preserve previously acquired capabilities.

\subsection{Parameter-space Exploration}
Most reinforcement learning methods for LLMs explore by sampling different trajectories from the same policy. Parameter-space exploration instead perturbs the model parameters, allowing different policies to produce different behaviours~\citep{sehnke2008policy,plappert2018parameter}. Evolution Strategies (ES) extend parameter-space exploration by evaluating a population of perturbed policies and using their fitness to update the model~\citep{salimans2017evolution}. Recent work shows that ES can scale to billion-parameter LLMs~\citep{qiu2026evolution}. \ours combines these two forms of exploration. It uses ES-style parameter perturbations to generate diverse policies, while their trajectories are converted into group-relative learning signals to update the student with an advantage-weighted objective.

\section{Method}
\label{sec:method}

\begin{figure}[t]  
    \centering
    \includegraphics[width=1\linewidth]{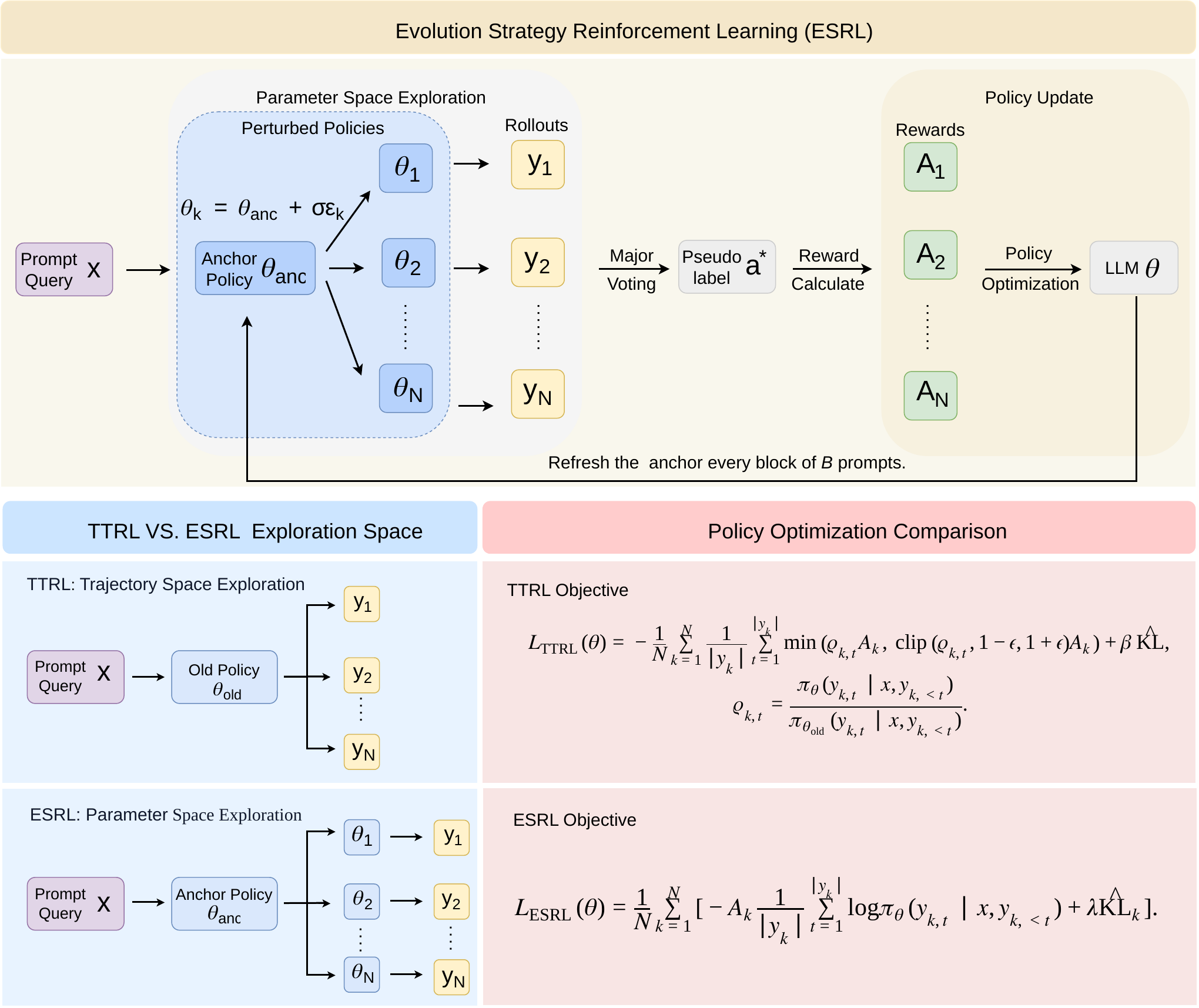}  
    \vspace{-0.2in}
    \caption{Overview of \ours. \ours first constructs a population of perturbed policies around a shared anchor and generates one rollout from each policy. Majority voting provides label-free rewards, which are converted into group-relative advantages. The student is then updated with an advantage-weighted log-likelihood objective. The anchor remains fixed within each block of $B$ prompts and is refreshed from the latest student at the next block boundary. }
    \label{fig:main_figure}
\end{figure}



In this section, we introduce \textbf{Evolution Strategy Reinforcement Learning (ESRL)}, a continual test-time learning framework. At a high level, \ours consists of two phases. In Phase 1 (\secref{sec:parameter-space}), \ours performs parameter-space population exploration and assigns rewards to the resulting rollouts. In Phase 2 (\secref{sec:update}), \ours converts these group-relative rewards into student updates using an advantage-weighted log-likelihood objective within each block.


\subsection{Parameter-space Population Exploration}
\label{sec:parameter-space}

Let $\theta$ denote the trainable parameters of the student. We draw $N$ independent perturbation directions
$\varepsilon_1,\dots,\varepsilon_N$, where
$\varepsilon_k\sim\mathcal{N}(0,I)$, and form $N$ perturbed policies:
\begin{equation}
  \theta_k
  \;=\;
  \theta_{\mathrm{anc}}+\sigma\varepsilon_k,
  \qquad k=1,\dots,N,
  \label{eq:perturb}
\end{equation}
where $\sigma$ is the perturbation radius. For the unlabeled prompt \(x\), rollout $k$ is then generated by its
corresponding policy:
\begin{equation}
  y_k
  \;\sim\;
  \pi_{\theta_k}(\,\cdot\mid x).
  \label{eq:rollout}
\end{equation}
Thus, the $N$ rollouts are generated by $N$ distinct perturbed policies rather than repeatedly sampled from a single policy. 


\paragraph{Label-free reward and group-relative advantage}
\label{sec:reward}
Given the rollout group $\{y_k\}_{k=1}^{N}$ for a prompt $x$, we extract
$a_k=\mathrm{Ans}(y_k)$ and choose the majority answer as the pseudo label:
\begin{equation}
  a^\star
  \;\in\;
  \operatorname*{arg\,max}_{a}
  \bigl|\{k:a_k=a\}\bigr|,
  \qquad
  r_k
\;=\;
\mathbf{1}\!\left[a_k=a^\star\right].
  \label{eq:reward}
\end{equation}
The rewards are standardized within the group to obtain the group-relative
advantages:
\begin{equation}
  A_k
  \;=\;
  \frac{r_k-\bar{r}}
       {\operatorname{Std}_{j=1,\dots,N}(r_j)+\epsilon},
  \qquad
  \bar{r}
  \;=\;
  \frac{1}{N}\sum_{j=1}^{N}r_j,
  \label{eq:advantage}
\end{equation}
where $\epsilon>0$ is a small numerical constant. We retain only rollout groups that contain a clear majority without complete agreement. The majority answer must be supported by at least two rollouts, while its fraction in the group must remain below $p_{\max}$.

\subsection{Group-Relative Student Update}
\label{sec:update}

For each retained prompt $x$ with rollout advantage pairs
$\{(y_k,A_k)\}_{k=1}^{N}$, we update the student using an
advantage-weighted log-likelihood objective:
\begin{equation}
  \mathcal{L}(\theta)
  \;=\;
  \frac{1}{N}
  \sum_{k=1}^{N}
  \Biggl[
    -A_k
    \underbrace{
      \frac{1}{|y_k|}
      \sum_{t=1}^{|y_k|}
      \log
      \pi_\theta
      \!\left(
        y_{k,t}\mid x,y_{k,<t}
      \right)
    }_{\text{mean per-token log-likelihood}}
    +
    \lambda\widehat{\mathrm{KL}}_k
  \Biggr],
  \label{eq:loss}
\end{equation}
where $\lambda$ controls regularisation toward the frozen initial policy
$\pi_{\theta_0}$. Positive advantages increase the likelihood of
consensus-supported trajectories, while negative advantages suppress
inconsistent ones. We use the non-negative $k_3$ estimator for the KL term:
\begin{equation}
  \begin{aligned}
    \widehat{\mathrm{KL}}_k
    &=
    \frac{1}{|y_k|}
    \sum_{t=1}^{|y_k|}
    \left(
      e^{\delta_{k,t}}
      -
      \delta_{k,t}
      -
      1
    \right),
    \\
    \delta_{k,t}
    &=
    \log\pi_{\theta_0}
    \!\left(y_{k,t}\mid x,y_{k,<t}\right)
    -
    \log\pi_\theta
    \!\left(y_{k,t}\mid x,y_{k,<t}\right).
  \end{aligned}
  \label{eq:kl}
\end{equation}

\paragraph{Block-wise anchoring}
The rollout population should remain centred on a stable policy while still
tracking the evolving student. Re-anchoring after every prompt would make
batched population generation inefficient, whereas keeping the same anchor
throughout training would cause the population to explore around an outdated
policy. We therefore process prompts in blocks of $B$ examples. At the
beginning of each block, we take a snapshot of the current student and use it
as the anchor to generate $N$ perturbed models. The anchor remains fixed within the block and is refreshed to the latest student at the next block boundary.

\paragraph{Memory-efficient perturbations.}
Within each block, the same $N$ perturbation directions are shared across all
$B$ prompts, allowing each population member to generate its $B$ trajectories
in a single batched call. To avoid materialising $N$ model copies, each
perturbation is represented by an integer seed $s_k$. For population member
$k$, we deterministically regenerate $\varepsilon(s_k)$ and temporarily set
the trainable parameters to
$\theta_{\mathrm{anc}}+\sigma\varepsilon(s_k)$ during generation. The
parameters are then restored to the anchor before moving to the next member.
This avoids storing $N$ perturbed model copies simultaneously. At any time, we keep only the anchor parameters and a single temporary perturbation buffer, while each population member is represented by a lightweight random seed.


\section{Experiment}

We evaluate our proposed method, \ours, under both single-task test-time adaptation (\secref{sub:main_results}) and multi-task continual learning (\secref{sub:continual_learning}). 

\subsection{Experimental Setup}
\paragraph{Models.}
To evaluate the generalization of ESRL across different backbone models, we conduct experiments using models of various scales from different families. We use Qwen3-0.6B,
Qwen3-4B,
Qwen3-8B~\cite{yang2025qwen3},
and Llama-3.2-3B~\cite{meta2024llama32}.

\paragraph{Baselines.}
We compare against the following representative methods: TTRL~\cite{zuo2025ttrl}, ECHO~\cite{zhao2026echo}, and EVOL-RL~\citep{zhou2025evolving}. We further include two additional baselines adapted to our setting: ES~\cite{qiu2026evolution} and SFT~\cite{ouyang2022training}. We report Pass@$n$ for all methods to ensure a consistent evaluation protocol. Additional baseline explanations and implementation details are provided in \appref{sec:experimental_setup}.

\paragraph{Benchmarks.}
We evaluate \ours across three domains. For each domain, we use one benchmark as the source task for adaptation and \sourcebox{source-benchmark evaluation}, and additional benchmarks from the same capability domain to measure \indomainbox{in-domain generalization}. We further use benchmarks from other capability domains to assess \outdomainbox{out-of-domain generalization}. 
For mathematics, we adapt the model on MATH-500~\cite{lightman2023verify} and evaluate on held-out MATH-500 examples, with AMC23~\cite{amc23} and AIME25~\cite{aime25} used to test in-domain generalization. 
For general knowledge, we adapt on MMLU~\cite{hendrycks2021mmlu} and evaluate on held-out MMLU examples, with ARC-Challenge~\cite{clark2018arc} and GPQA-Diamond~\cite{rein2024gpqa} as in-domain generalization benchmarks. 
For the medical domain, we adapt on the MedQA-USMLE training set~\cite{jin2021medqa} and evaluate on the MedQA test set, with MedMCQA~\cite{pal2022medmcqa} and PubMedQA~\cite{jin2019pubmedqa} used for in-domain generalization.

\vspace{-0.10in}
\subsection{Main Results}
\label{sub:main_results}
\tabref{tab:main_results} summarizes the main results across different model backbones and domains. Across all four backbone models, \textbf{\ours consistently outperforms the baselines in average accuracy.} The largest relative gains reach 31.65\% over the base model and 16.61\% over the second-best method. The gains are also consistent across training domains and model scales: \ours improves upon or matches the base model on 35 of 36 backbone--benchmark combinations and achieves the best or tied-best performance on 29 of 36. Additionally, \textbf{the improvement extends beyond the source benchmarks.} Averaged across the four backbones, \ours improves source-benchmark accuracy from 44.33 to 52.90 and in-domain generalization accuracy from 39.60 to 47.40, corresponding to relative gains of 19.32\% and 19.71\%, respectively. The gains are particularly strong in the Math domain, where average accuracy increases from 17.59 to 31.40, with \ours achieving the best or tied-best result on all 12 Math evaluations.

\begin{table*}[t]
\centering
\caption{
Pass@1 results for source-benchmark evaluation and in-domain generalization across different backbone models.
}
\setlength{\tabcolsep}{4pt}
\resizebox{\textwidth}{!}{
\begin{tabular}{llBGG|BGG|BGG|c}
\toprule

\multirow{2}{*}{\textbf{Model}} 
& \multirow{2}{*}{\textbf{Method}}
& \multicolumn{3}{c|}{\textbf{Math}}
& \multicolumn{3}{c|}{\textbf{General Knowledge}}
& \multicolumn{3}{c|}{\textbf{Medical}}
& \multirow{2}{*}{\textbf{Avg.}} \\

\cmidrule(lr){3-5}
\cmidrule(lr){6-8}
\cmidrule(lr){9-11}

&
& \textbf{MATH-500} 
& \textbf{AMC23} 
& \textbf{AIME25}
& \textbf{MMLU} 
& \textbf{ARC-C} 
& \textbf{GPQA-D}
& \textbf{MedQA} 
& \textbf{MedMCQA} 
& \textbf{PubMedQA}
& \\

\midrule


\multirow{7}{*}{Qwen3-0.6B}
& Base model      
& 5.0 
& 6.0 
& 0.0 
& 44.0 
& 65.5 
& 26.3 
& 23.5 
& 37.0 
& 45.5 
& 28.1 \\

& TTRL            
& 10.0 
& 13.3 
& \underline{3.3} 
& 44.5 
& \underline{70.0} 
& 25.8 
& \underline{32.5} 
& 33.5 
& 52.5 
& \underline{31.7} \\

& ES              
& 6.0 
& 6.0 
& 0.0 
& \underline{47.0} 
& 68.5 
& 27.3 
& 28.0 
& \underline{37.5} 
& 45.5 
& 29.5 \\

& SFT             
& 0.0 
& 0.0 
& 0.0 
& 43.0 
& 60.5 
& \underline{28.3} 
& 22.5 
& 23.0 
& \textbf{60.0} 
& 26.4 \\

& ECHO            
& \underline{17.0} 
& 9.9 
& \underline{3.3} 
& 46.0 
& 69.5 
& 23.7 
& 18.5 
& \underline{37.5} 
& 45.5 
& 30.1 \\

& EVOL-RL         
& 4.0 
& \underline{14.9} 
& 0.0 
& 46.5 
& 69.5 
& 27.3 
& 22.5 
& 32.0 
& 56.0 
& 30.3 \\

& \textbf{ESRL}   
& \textbf{27.0} 
& \textbf{19.3} 
& \textbf{6.7} 
& \textbf{49.5} 
& \textbf{70.5} 
& \textbf{30.8} 
& \textbf{34.0} 
& \textbf{38.0} 
& \underline{57.0} 
& \textbf{37.0} \\

\midrule


\multirow{7}{*}{Qwen3-4B}
& Base model      
& 55.0 
& 24.1 
& 10.0 
& 69.0 
& 88.0 
& 33.8 
& \underline{52.0} 
& 55.5 
& 67.0 
& 50.5 \\

& TTRL            
& 70.0 
& 45.8 
& \underline{13.3} 
& 67.5 
& \underline{90.0} 
& 35.4 
& 51.0 
& \textbf{58.0} 
& 62.5 
& 54.8 \\

& ES              
& 53.0 
& 30.1 
& 3.3 
& 68.0 
& 88.0 
& 34.8 
& 50.0 
& 55.5 
& 66.0 
& 49.9 \\

& SFT             
& 8.0 
& 10.3 
& 0.0 
& 60.0 
& 84.5 
& 26.8 
& 50.5 
& \textbf{58.0} 
& \underline{68.0} 
& 40.7 \\

& ECHO            
& \underline{71.0} 
& 47.7 
& \underline{13.3} 
& \underline{70.0} 
& 89.5 
& 34.3 
& 51.0 
& \underline{56.5} 
& 62.5 
& \underline{55.1} \\

& EVOL-RL         
& 68.0 
& \underline{48.2} 
& 3.3 
& 65.5 
& 89.5 
& \textbf{36.4} 
& 50.0 
& 56.0 
& 65.0 
& 53.5 \\

& \textbf{ESRL}   
& \textbf{74.0} 
& \textbf{51.8} 
& \textbf{16.7} 
& \textbf{71.5} 
& \textbf{98.0} 
& \underline{35.9} 
& \textbf{58.5} 
& \textbf{58.0} 
& \textbf{69.0} 
& \textbf{59.3} \\

\midrule


\multirow{7}{*}{Qwen3-8B}
& Base model      
& 65.0 
& 27.3 
& \underline{13.3} 
& \textbf{73.5} 
& 90.5 
& 36.9 
& 57.5 
& 60.0 
& \underline{71.5} 
& 55.1 \\

& TTRL            
& \underline{72.0} 
& \underline{55.4} 
& \textbf{16.7} 
& 69.5 
& 92.0 
& \underline{40.9} 
& \textbf{62.0} 
& 61.5 
& \underline{71.5} 
& \underline{60.2} \\

& ES              
& 61.0 
& 20.5 
& 3.3 
& \textbf{73.5} 
& 91.0 
& 37.4 
& 58.5 
& 60.0 
& 70.5 
& 52.9 \\

& SFT             
& 13.0 
& 2.4 
& 0.0 
& 67.5 
& 89.5 
& 35.4 
& 58.5 
& \textbf{62.5} 
& \underline{71.5} 
& 44.5 \\

& ECHO            
& \underline{72.0} 
& 45.8 
& 10.0 
& 72.0 
& 91.5 
& 40.4 
& 58.5 
& \textbf{62.5} 
& \underline{71.5} 
& 58.2 \\

& EVOL-RL         
& 71.0 
& \underline{55.4} 
& \underline{13.3} 
& 71.0 
& \underline{92.5} 
& 38.4 
& \underline{60.5} 
& \underline{62.0} 
& \underline{71.5} 
& 59.5 \\

& \textbf{ESRL}   
& \textbf{78.0} 
& \textbf{57.5} 
& \textbf{16.7} 
& \underline{72.5} 
& \textbf{100.0} 
& \textbf{41.4} 
& \underline{60.5} 
& \textbf{62.5} 
& \textbf{72.5} 
& \textbf{62.4} \\

\midrule


\multirow{7}{*}{Llama-3.2-3B}
& Base model      
& 3.0 
& 2.4 
& 0.0 
& 44.5 
& 55.0 
& \textbf{29.8} 
& 40.0 
& 43.5 
& 61.5 
& 31.1 \\

& TTRL            
& \underline{7.0} 
& 3.2 
& \textbf{3.3} 
& 53.0 
& \textbf{71.0} 
& 24.2 
& 44.0 
& 49.0 
& \underline{65.0} 
& 35.5 \\

& ES              
& 0.0 
& 0.0 
& 0.0 
& 43.5 
& 55.8 
& 27.8 
& 41.5 
& 46.0 
& 62.5 
& 30.8 \\

& SFT             
& 0.0 
& 0.0 
& 0.0 
& 20.5 
& 27.0 
& 25.8 
& 12.5 
& 19.5 
& 53.0 
& 17.6 \\

& ECHO            
& 4.0 
& \underline{3.6} 
& 0.0 
& \textbf{55.5} 
& \underline{69.0} 
& \underline{28.3} 
& \textbf{46.0} 
& \underline{52.0} 
& \textbf{65.5} 
& \underline{36.0} \\

& EVOL-RL         
& 3.2 
& 1.2 
& 0.0 
& 53.0 
& \underline{69.0} 
& 26.3 
& 43.5 
& 49.5 
& 63.0 
& 34.3 \\

& \textbf{ESRL}   
& \textbf{11.0} 
& \textbf{14.8} 
& \textbf{3.3} 
& \underline{53.5} 
& 68.5 
& \textbf{29.8} 
& \underline{44.8} 
& \textbf{53.5} 
& \textbf{65.5} 
& \textbf{38.3} \\

\bottomrule
\end{tabular}
}
\label{tab:main_results}
\vspace{-0.15in}
\end{table*}

\vspace{-0.10in}
\subsection{Continual Test-Time Learning}
\label{sub:continual_learning}

In this section, we investigate a long-horizon continual learning setting to examine whether the degradation observed above accumulates over sequential adaptation. Following the protocol of \citet{shenfeld2026self}, a single model is trained on three tasks in sequence, while all tasks are evaluated continuously throughout training.The model is trained sequentially on MATH-500~\citep{lightman2023verify},
MMLU~\citep{hendrycks2021mmlu}, and MedQA~\citep{jin2021medqa}.

\begin{figure}[t]  
    \centering
    \includegraphics[width=1\linewidth]{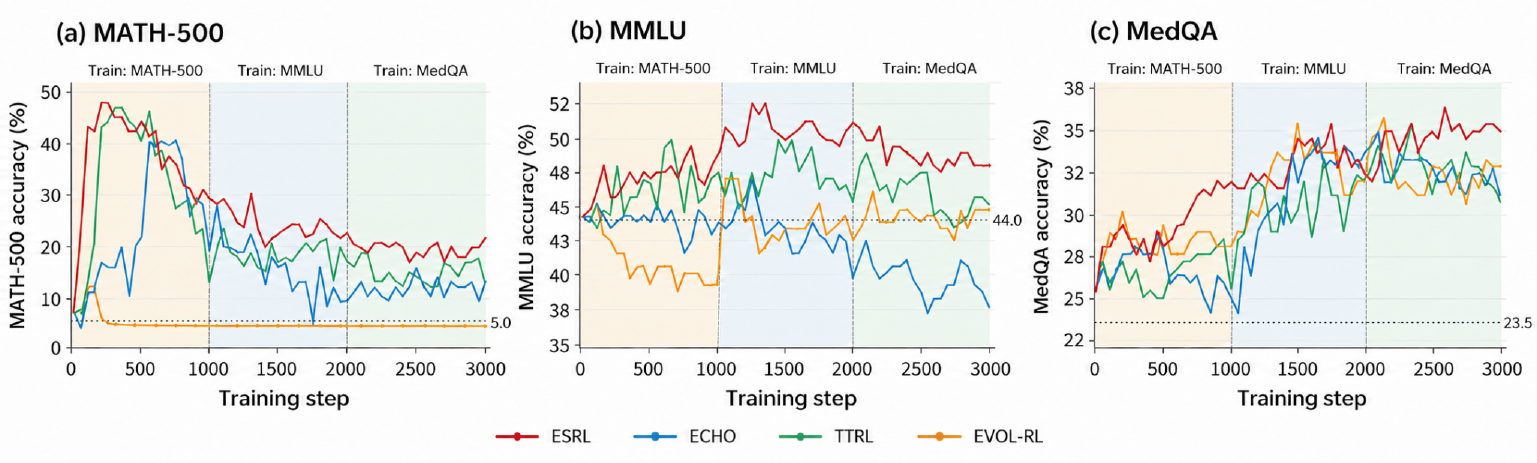}  
    \vspace{-0.2in}
    \caption{Sequential training of Qwen3-0.6B on MATH-500, MMLU, and MedQA. Pass@1 is reported across 3000 training steps for (a) MATH-500, (b) MMLU, and (c) MedQA.
}
    \label{fig:sequential_figure}
    \vspace{-0.0in}
\end{figure}

As shown in Figure~\ref{fig:sequential_figure}, \ours progressively accumulates new capabilities while better preserving those acquired earlier in the sequence. On MATH-500, which is learned first, \ours shows substantially slower degradation after training shifts to MMLU and MedQA. On MMLU, \ours improves during the second training stage and retains most of this gain after the model moves to MedQA, whereas the baselines exhibit larger degradation. Finally, on MedQA, \ours continues to acquire the new capability during the last stage and achieves the strongest final performance. Together, these results show that \ours better balances the acquisition of new capabilities with the retention of previously learned ones during long-horizon sequential adaptation.

\section{Analysis}
\begin{figure}[t]  
    \centering
    \includegraphics[width=1\linewidth]{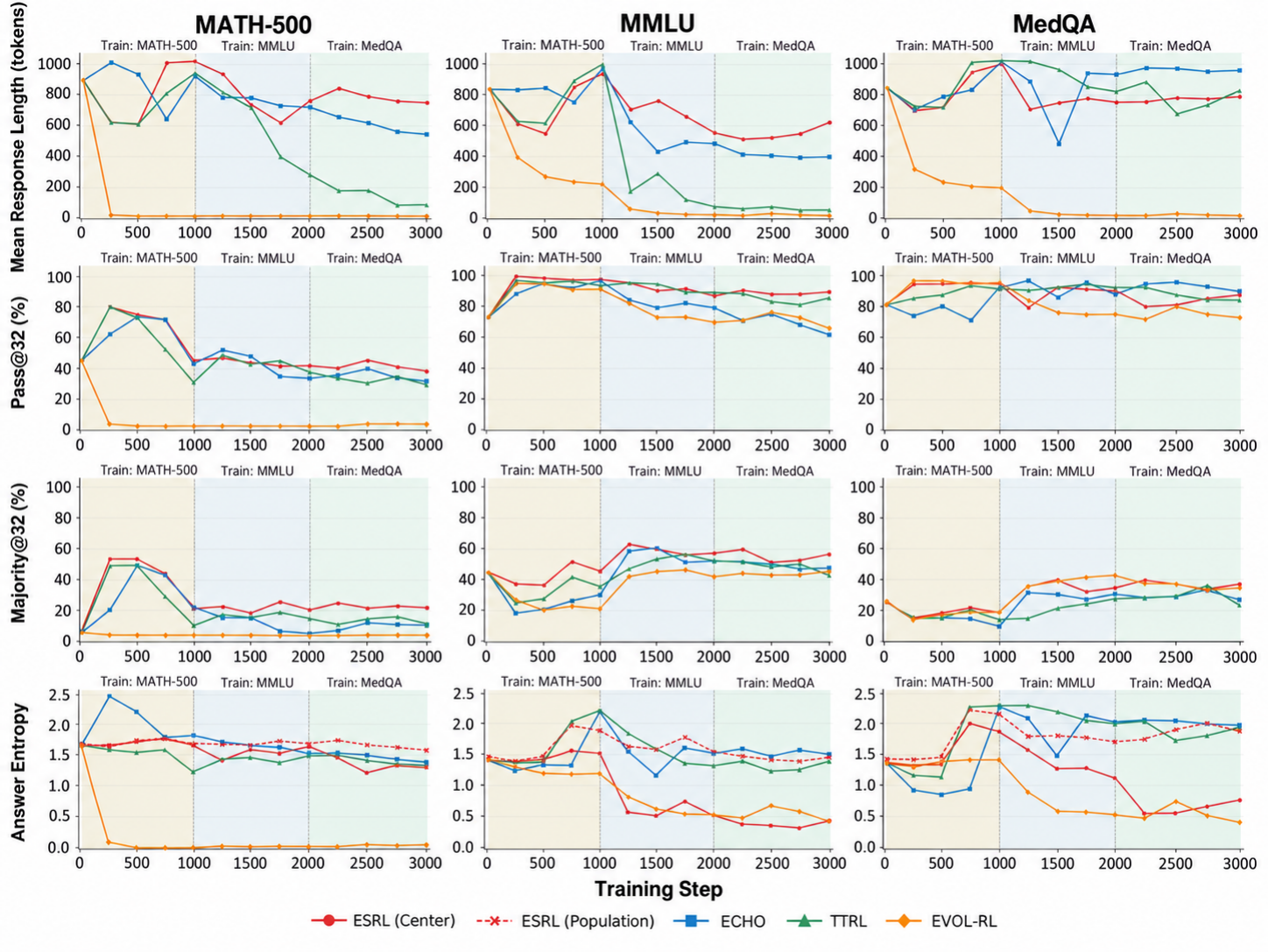}  
    \vspace{-0.2in}
    \caption{Training dynamics during continual learning. The first row shows the mean response length, the second row shows Pass@32, the third row shows Majority@32, where \ours reports population-level majority voting, and the fourth row shows answer entropy. The three columns correspond to MATH-500, MMLU, and MedQA, respectively. All metrics are evaluated every 250 training steps. Detailed definitions of these metrics are provided in \appref{sec:evaluation_metrics}
}
    \label{fig:anaysis_figure}
    
\end{figure}

\subsection{Why Policy-Level Exploration Works}
To better understand the mechanism underlying \ours, we investigate why policy-level exploration can better mitigate catastrophic forgetting and have better performance in continual learning tasks than trajectory-level exploration methods.

\paragraph{Takeaway 1: \ours concentrates the center policy without policy collapse.}
The results show clear differences in how the compared methods shape their output distributions. EVOL-RL has very low answer entropy, suggesting that its output distribution becomes highly concentrated. However, its response length also drops sharply during training, together with a clear decrease in Pass@32.
This pattern suggests policy collapse: the model becomes concentrated on a small set of outputs and increasingly confined to a narrow local region of the reasoning space, reducing its ability to explore alternative reasoning modes during subsequent tasks and resulting in worse continual-learning performance.

TTRL and ECHO show the opposite behavior. Their relatively high answer entropy indicates a more diffuse output distribution, while high Pass@32 suggests that they preserve broad reasoning coverage, so correct solutions can still be found when multiple rollouts are sampled.  However, the much lower Pass@1 suggests that correct solutions are often assigned relatively low probability. As a result, the model may reach the correct answer after several rollouts, while being much less likely to produce it in a single attempt. This is less useful in verifier-free settings, where generating more candidates does not by itself tell us which one is correct.

ESRL avoids both extremes. Its center policy has lower answer entropy than TTRL and ECHO, while its response length does not show the sharp contraction observed in EVOL-RL.
This suggests that ESRL concentrates probability mass on a smaller set of useful outputs without collapsing the center policy, which is also consistent with its stronger Pass@1.

\paragraph{Takeaway 2: Population exploration preserves broad reasoning coverage.}

ESRL maintains broader exploration through its population of locally perturbed policies.
As shown in the last row of Fig.~\ref{fig:anaysis_figure}, ESRL (Population) maintains substantially higher answer entropy than the center policy, indicating that the perturbed policies cover a broader set of outputs.
The higher population entropy suggests that ESRL preserves a broader reasoning space during continual learning. This broader coverage makes the model less likely to become trapped in a narrow local region of the reasoning space, helping preserve alternative reasoning modes across tasks and alleviate catastrophic forgetting.

\paragraph{Takeaway 3: Population consensus provides a reliable learning signal.}
As shown in Fig.~\ref{fig:anaysis_figure}, \ours Population-Maj@32 remains relatively high, particularly in the later stages of training. This shows that, despite broader exploration, the perturbed population can still form a reliable consensus around the correct answer. Since majority voting is used to construct the learning signal, this provides ESRL with effective supervision without an external verifier. This becomes especially important in later stages of continual learning, where population-level exploration keeps alternative reasoning modes reachable, while a high Maj@32 ensures that these diverse rollouts can still produce a useful dominant signal for continued optimization.

\subsection{Useful Stochasticity vs. Signal Dilution}

\begin{wrapfigure}{r}{0.5\textwidth}
\centering

\includegraphics[width=\linewidth]{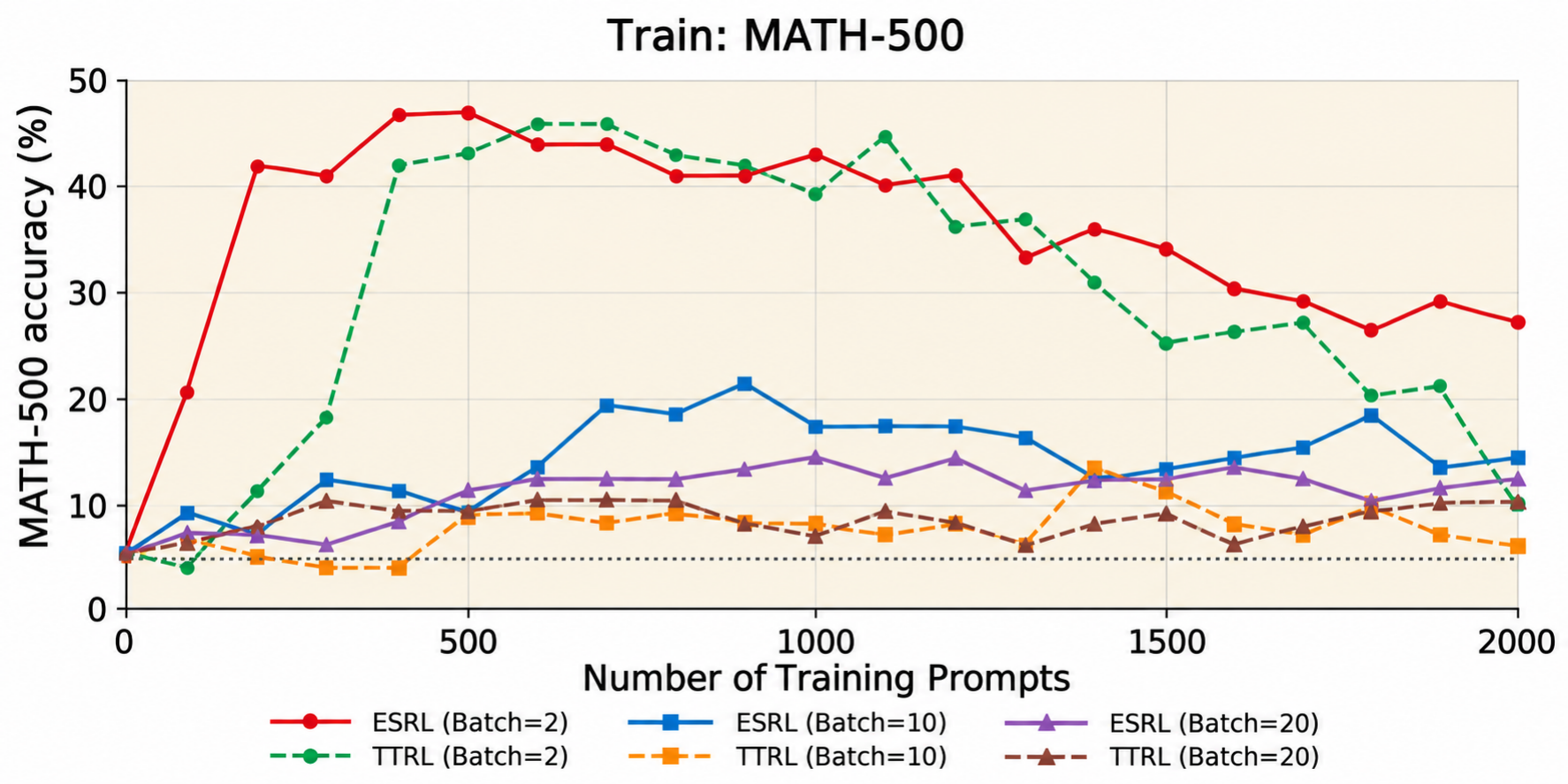}

\vspace{-0.15in}
\caption{Comparison of Pass@1 on MATH-500 under different batch sizes.}
\label{fig:batch_size_compare}
\vspace{-0.15in}
\end{wrapfigure}

We further investigate why MATH-500 accuracy first improves rapidly and then declines during the MATH training stage of continual test-time learning. This behavior is closely related to how self-generated signals are aggregated. Since continual test-time learning has no ground-truth labels, supervision comes entirely from the model's own rollouts. Some of these signals can reveal useful reasoning modes that are not yet captured by the current policy. With a small batch, such sparse informative signals have a larger influence on each update, enabling rapid early improvement. As the batch size increases, these signals are averaged with more weak or conflicting updates, leading to more stable but weaker adaptation. However, the higher variance of small batches also makes them more sensitive to noisy pseudo-labels. As training continues and useful signals become less frequent, noise can gradually dominate the updates and cause performance to decline. As shown in \figref{fig:batch_size_compare}, with a batch size of 2, MATH-500 accuracy quickly rises to around 45\% before gradually decreasing to 27\%. A batch size of 10 shows a similar pattern, but with a lower peak of around 20\%, while a batch size of 20 produces a much smoother increase and reaches only around 10\%. TTRL shows a similar batch-size-dependent pattern. Overall, batch size plays an important role in continual test-time learning. Small batches preserve sparse useful exploratory signals and enable faster adaptation, but are more sensitive to noise, while larger batches provide more stable updates at the cost of diluting informative signals.

\section{Ablation Study}

\paragraph{Block Size}
We further study the effect of the block size $B$ on Qwen3-0.6B across all three domains. As shown in Table~\ref{tab:block_size}, performance is relatively robust to $B$, with the default setting $B=10$ achieving or matching the best result on six of the nine benchmarks. Performance degrades at $B=50$, as the anchor remains fixed while the student continues to update. This causes the population to explore around an increasingly stale policy and increases the mismatch between the generated trajectories and the current student. When $B=2$, the anchor is refreshed after every training step. However, this reduces batching efficiency because each perturbed policy processes only a few prompts before the population is reconstructed. Overall, a moderate block size provides a better balance between computational efficiency, exploration stability, and alignment with the evolving student.

\begin{table*}[htp]
\vspace{-0.2in}
\centering
\small
\caption{Pass@1 under different block sizes $B$ using Qwen3-0.6B.}
\label{tab:block_size}
\setlength{\tabcolsep}{6pt}
\renewcommand{\arraystretch}{1.08}

\resizebox{\textwidth}{!}{%
\begin{tabular}{@{}cBGGBGGBGGc@{}}
\toprule

\textbf{$B$}
& \multicolumn{3}{c}{\textbf{Math}}
& \multicolumn{3}{c}{\textbf{General Knowledge}}
& \multicolumn{3}{c}{\textbf{Medical}}
& \textbf{Avg.} \\

\cmidrule(lr){2-4}
\cmidrule(lr){5-7}
\cmidrule(lr){8-10}

& \textbf{MATH-500}
& \textbf{AMC23}
& \textbf{AIME25}
& \textbf{MMLU}
& \textbf{ARC-C}
& \textbf{GPQA-D}
& \textbf{MedQA}
& \textbf{MedMCQA}
& \textbf{PubMedQA}
& \\

\midrule

2
& 23.0
& 16.9
& \textbf{6.7}
& 48.0
& 67.0
& 30.3
& 31.0
& \textbf{41.0}
& 48.0
& 34.7 \\

\textbf{10}
& \textbf{27.0}
& 19.3
& \textbf{6.7}
& \textbf{49.5}
& \textbf{70.5}
& 30.8
& \textbf{34.0}
& 38.0
& \textbf{57.0}
& \textbf{37.0} \\

20
& 25.0
& 15.3
& \textbf{6.7}
& \textbf{49.5}
& 65.5
& \textbf{32.3}
& 30.5
& 35.0
& 54.5
& 34.9 \\

50
& 17.0
& \textbf{22.9}
& 0.0
& 46.0
& 63.5
& 27.3
& \textbf{34.0}
& 39.0
& 47.5
& 33.0 \\

\bottomrule
\end{tabular}%
}
\end{table*}

\paragraph{Group size.}

\begin{wraptable}{r}{0.5\textwidth}
\vspace{-0.3in}
\centering
\small
\caption{Pass@1 under different rollout group sizes $N$ using Qwen3-4B.}
\vspace{-0.00in}

\setlength{\tabcolsep}{4pt}

\resizebox{\linewidth}{!}{%
\begin{tabular}{cBGGc}
\toprule

\multirow{2}{*}{\textbf{Group Size $N$}}
& \multicolumn{3}{c}{\textbf{Math}}
& \multirow{2}{*}{\textbf{Avg.}} \\

\cmidrule(lr){2-4}

& \textbf{MATH-500}
& \textbf{AMC23}
& \textbf{AIME25}
& \\

\midrule

16
& 74.0
& 51.8
& \textbf{16.7}
& 47.5 \\

32
& 72.0
& 50.6
& \textbf{16.7}
& 46.4 \\

64
& \textbf{75.0}
& \textbf{53.0}
& \textbf{16.7}
& \textbf{48.2} \\

\bottomrule
\end{tabular}%
}

\label{tab:group_size}
\vspace{-0.1in}
\end{wraptable}
We further study the effect of the rollout group size $N$ on the Math domain using Qwen3-4B. We compare $N\in\{16,32,64\}$ while keeping all other settings unchanged. As shown in \tabref{tab:group_size}, the performance of \ours remains relatively stable across different group sizes, indicating that the method is robust to this hyperparameter. A larger group size can provide broader exploration and a more reliable group-level learning signal. However, it can also introduce more noise into the pseudo-labeling process, so increasing $N$ does not necessarily lead to better performance.

\paragraph{Loss function.}
We investigate how population-generated trajectories should be used to update the student. We keep the policy-level exploration and rollout generation of \ours unchanged, and compare our advantage-weighted log-likelihood objective with a GRPO-style objective. Since each rollout in \ours is generated by its own perturbed policy, the GRPO variant uses the corresponding $\pi_{\theta_k}$ in the importance ratio. The two variants therefore use the same rollouts and group-relative advantages, differing only in the student update: our objective directly optimizes trajectory likelihood, whereas GRPO uses a clipped importance-ratio objective. The advantage-weighted loss performs better across both in-domain and out-of-domain evaluation. Averaged over the three Math benchmarks, performance improves from 42.8 to 47.5, while the average across the six out-of-domain benchmarks increases from 48.9 to 52.1. It improves on eight of the nine benchmarks and matches GRPO on AIME25, suggesting that direct advantage-weighted likelihood optimization is better suited to the population-generated trajectories in \ours.

\begin{table*}[htp]
\vspace{-0.2in}
\centering
\small
\caption{Pass@1 under different optimization objectives using Qwen3-4B.}
\label{tab:loss_function}
\setlength{\tabcolsep}{4pt}
\renewcommand{\arraystretch}{1.08}

\resizebox{\textwidth}{!}{%
\begin{tabular}{@{}lBGGRRRRRRc@{}}
\toprule

\textbf{Loss Function}
& \multicolumn{3}{c}{\textbf{Math}}
& \multicolumn{3}{c}{\textbf{General Knowledge}}
& \multicolumn{3}{c}{\textbf{Medical}}
& \textbf{Avg.} \\

\cmidrule(lr){2-4}
\cmidrule(lr){5-7}
\cmidrule(lr){8-10}

& \textbf{MATH-500}
& \textbf{AMC23}
& \textbf{AIME25}
& \textbf{MMLU}
& \textbf{ARC-C}
& \textbf{GPQA-D}
& \textbf{MedQA}
& \textbf{MedMCQA}
& \textbf{PubMedQA}
& \\

\midrule

GRPO-Style Clipped Loss
& 67.0
& 44.6
& \textbf{16.7}
& 65.5
& 87.5
& 29.3
& 49.0
& 54.0
& 61.5
& 52.8 \\

Advantage-Weighted Loss
& \textbf{74.0}
& \textbf{51.8}
& \textbf{16.7}
& \textbf{68.5}
& \textbf{89.0}
& \textbf{33.3}
& \textbf{53.0}
& \textbf{56.0}
& \textbf{66.0}
& \textbf{56.5} \\

\bottomrule
\end{tabular}%
}
\end{table*}

\section{conclusion and Limitation}
In this paper, we introduced \ours, a continual test-time learning framework that combines parameter-space population exploration with label-free group-relative learning. By maintaining a population of locally perturbed policies around an evolving anchor, \ours enables broader exploration while preserving a stable center policy for continual adaptation. Experiments across diverse reasoning domains show that \ours consistently improves test-time learning and better retains previously acquired capabilities during long-horizon continual learning. Further analysis suggests that its advantage comes from maintaining broad reasoning coverage without collapsing the center policy. Overall, our results highlight policy-level population exploration as a promising approach to continual test-time adaptation without external supervision.
\paragraph{Limitation.} Although \ours is evaluated across four backbone models and nine benchmarks, broader  validation on more diverse  and high-stakes tasks would further strengthen its empirical robustness. We leave this direction for future work. 


\clearpage

\clearpage

\bibliographystyle{iclr2027_conference}
\bibliography{iclr2027_conference}


\clearpage
\appendix

\section{Experimental Setup}
\label{sec:experimental_setup}

\subsection{Experiment Details}
In this section, we provide detailed descriptions of our baselines implemented in our experiment.

\paragraph{TTRL.}
TTRL~\cite{zuo2025ttrl} performs reinforcement learning directly on unlabeled test-time data. For each prompt, it samples multiple responses from the current policy and uses majority voting over the extracted answers to obtain a pseudo-label. Rollouts that agree with the pseudo-label receive positive rewards, which are then used for policy optimization.

\paragraph{ECHO.}
ECHO~\cite{zhao2026echo} extends test-time reinforcement learning by improving both rollout exploration and policy optimization. It combines entropy--confidence guided rollout generation with confidence-aware advantage shaping to reduce early pseudo-label bias and rollout collapse.

\paragraph{EVOL-RL.}
EVOL-RL~\cite{zhou2025evolving} augments majority-based test-time reinforcement learning with an explicit novelty signal. In addition to rewarding agreement with the majority answer, it encourages semantically distinct trajectories to maintain reasoning diversity and reduce entropy collapse.

\paragraph{ES.}
ES~\cite{qiu2026evolution} performs zeroth-order optimization directly in parameter space. It samples a population of perturbed policies, evaluates each member using a scalar fitness value, and aggregates their fitness-weighted perturbations to update the model. We adapt ES to our test-time setting by applying it to the same unlabeled prompts and using the corresponding self-generated reward signal for population evaluation.

\paragraph{SFT.}
Supervised fine-tuning (SFT)~\cite{ouyang2022training} directly increases the likelihood of target responses using a token-level cross-entropy objective. Since ground-truth answers are unavailable in our setting, we adapt SFT with self-generated pseudo-targets from the test-time data. The resulting baseline therefore uses the same unlabeled inputs as \ours.

\subsection{Evaluation Metrics}
\label{sec:evaluation_metrics}

Let $Q$ denote the number of evaluation questions and $n$ the number of sampled
responses per question. For each question $x_i$, let
$\mathcal{Y}_i=\{y_{i,j}\}_{j=1}^{n}$ denote its $n$ sampled responses,
$A(y)$ the normalized final answer extracted from response $y$, and $a_i^*$
the ground-truth answer.

We consider two evaluation modes. In the \emph{center-policy} mode, all $n$
responses are sampled from the current student policy $\pi_\theta$. In the
\emph{population-policy} mode, responses are pooled across the ES-perturbed
policies. Unless otherwise stated, we use $n=32$, corresponding to 32 responses
from the center policy or two responses from each of the $N=16$ perturbed
policies.

\paragraph{Pass@$n$.}
Pass@$n$ measures whether at least one of the $n$ sampled responses produces
the correct final answer:
\[
\operatorname{Pass@}n
=
\frac{1}{Q}
\sum_{i=1}^{Q}
\mathbf{1}
\left[
\exists\, j\in\{1,\ldots,n\}:
A(y_{i,j})=a_i^*
\right].
\]
We report Pass@$1$ and Pass@$32$. Pass@$1$ measures single-sample performance,
while Pass@$32$ measures whether correct solutions remain accessible under a
larger sampling budget, providing a measure of reasoning coverage.

\paragraph{Majority@$n$.}
For each question, we select the most frequent answer among the $n$ sampled
responses:
\[
\hat a_i
=
\arg\max_a
\sum_{j=1}^{n}
\mathbf{1}
\left[
A(y_{i,j})=a
\right],
\]
and compute
\[
\operatorname{Maj@}n
=
\frac{1}{Q}
\sum_{i=1}^{Q}
\mathbf{1}
\left[
\hat a_i=a_i^*
\right].
\]
Under center-policy evaluation, Majority@$n$ measures whether repeated
sampling from the student forms a correct consensus. Under population-policy
evaluation, it measures whether the perturbed-policy population collectively
forms a correct consensus. Unparseable responses abstain from voting, and ties
are resolved by the first observed tied answer.

\paragraph{Answer Entropy.}
For question $i$, let $n_{i,a}$ denote the number of parseable responses among
the $n$ samples whose normalized answer is $a$. We define
\[
p_i(a)
=
\frac{n_{i,a}}
{\sum_{a'} n_{i,a'}},
\]
and compute
\[
H_i
=
-\sum_a p_i(a)\log p_i(a),
\qquad
\overline{H}
=
\frac{1}{Q}
\sum_{i=1}^{Q}H_i.
\]
Unparseable responses are excluded from the entropy calculation. Under
center-policy evaluation, answer entropy captures stochastic diversity within
a single policy. Under population-policy evaluation, it additionally captures
diversity across different perturbed policies introduced by parameter-space
exploration.

\paragraph{Mean Response Length.}
Let $\ell(y)$ denote the number of generated completion tokens in response $y$.
We compute
\[
\overline{L}
=
\frac{1}{nQ}
\sum_{i=1}^{Q}
\sum_{j=1}^{n}
\ell(y_{i,j}).
\]
We use this metric to track changes in reasoning length during continual adaptation.

\subsection{Implementation Details}
\label{sec:implementation_details}

For hyperparameters, we use the AdamW optimizer with \texttt{bfloat16} precision and apply gradient clipping with a maximum norm of $1.0$. For rollout generation, we use a default group/population size of 16, with each population member generating one response. Responses are sampled with a temperature of $0.8$ and top-$p$ of $0.95$. For optimization, we use a learning rate of $1\times10^{-5}$ for TTRL, ECHO, EVOL-RL, and \ours, while SFT uses $5\times10^{-5}$ and ES uses a step size of $5\times10^{-4}$. The KL coefficient is set to $1\times10^{-3}$, and the PPO clipping parameter is $0.2$. For \ours, we set the perturbation standard deviation to $\sigma=1\times10^{-3}$ and refresh the anchor every 10 prompts. Unless otherwise specified, the batch size is set to 2 for all methods, and $p_{\max}=0.95$. We set the number of episodes to 5, 4, and 4 for MATH-500, MMLU, and MedQA, respectively, according to the dataset size. The maximum generation length is set to 8,192 tokens for all methods. All experiments are conducted on 8$\times$ NVIDIA H200 141 GB GPUs.

\newpage
\section{Additional Experiment and Ablations results}

\subsection{Multi-Sample Evaluations}
\tabref{tab:pass32_results} presents the Pass@32 results for source-benchmark evaluation and in-domain generalization. The results are consistent with the main findings: \ours demonstrates strong generalization, achieving the best average performance and ranking first or second on 8 out of 9 benchmarks.

\begin{table*}[h]
\vspace{-0.2in}
\centering
\small
\caption{Pass@32 results for source-benchmark evaluation and in-domain generalization using Qwen3-0.6B.}
\label{tab:pass32_results}

\setlength{\tabcolsep}{4pt}
\renewcommand{\arraystretch}{1.05}

\resizebox{\textwidth}{!}{%
\begin{tabular}{lBGGBGGBGGc}
\toprule
& \multicolumn{3}{c}{\textbf{Math}}
& \multicolumn{3}{c}{\textbf{General Knowledge}}
& \multicolumn{3}{c}{\textbf{Medical}}
& \\

\cmidrule(lr){2-4}
\cmidrule(lr){5-7}
\cmidrule(lr){8-10}

\textbf{Method}
& \textbf{MATH-500}
& \textbf{AMC23}
& \textbf{AIME25}
& \textbf{MMLU}
& \textbf{ARC-C}
& \textbf{GPQA-D}
& \textbf{MedQA}
& \textbf{MedMCQA}
& \textbf{PubMedQA}
& \textbf{Avg.} \\
\midrule

Base model
& \textbf{45.0}
& 33.0
& \underline{6.7}
& 73.0
& \underline{85.0}
& \textbf{63.9}
& 81.0
& \underline{83.0}
& \textbf{90.7}
& \underline{62.4} \\

TTRL
& 31.0
& 36.1
& \underline{6.7}
& 75.0
& \textbf{92.0}
& 54.0
& \textbf{90.0}
& 73.0
& \underline{90.0}
& 60.9 \\

EVOL-RL
& 4.0
& \textbf{41.0}
& 0.0
& \textbf{86.0}
& \textbf{92.0}
& \underline{59.1}
& 86.0
& 68.0
& 84.0
& 57.8 \\

ECHO
& 43.0
& 37.3
& 3.6
& 62.0
& 79.0
& 34.8
& \underline{88.0}
& 72.0
& 87.0
& 56.3 \\

\ours
& \textbf{45.0}
& \underline{38.6}
& \textbf{13.3}
& \underline{82.0}
& \textbf{92.0}
& 57.4
& \underline{88.0}
& \textbf{88.0}
& \underline{90.0}
& \textbf{66.0} \\

\bottomrule
\end{tabular}%
}

\vspace{-0.1in}
\end{table*}

\subsection{Effect of Noise Scale on Performance}

\tabref{tab:sigma_ablation} reports the Pass@1 performance on the general knowledge domain under different values of $\sigma$. The default setting, $\sigma=0.001$, achieves the best performance on two of the three benchmarks and the highest average score. These results also show that \ours remains robust across different choices of $\sigma$. Notably, setting $\sigma=0$, corresponding to \ours w/o Parameter Perturbation, removes parameter-space exploration while keeping the remaining ESRL pipeline unchanged, reducing the average accuracy from 50.3 to 46.2 and confirming the benefit of parameter-space exploration.

\begin{table}[h]
\vspace{-0.2in}
\centering
\footnotesize
\caption{Pass@1 results under different perturbation scale $\sigma$ using Qwen3-0.6B.}
\label{tab:sigma_ablation}

\setlength{\tabcolsep}{3pt}
\renewcommand{\arraystretch}{1.02}

\begin{tabular}{cBGGc}
\toprule

\multirow{2}{*}{\textbf{$\sigma$}}
& \multicolumn{3}{c}{\textbf{General Knowledge}}
& \multirow{2}{*}{\textbf{Avg.}} \\

\cmidrule(lr){2-4}

& \textbf{MMLU}
& \textbf{ARC-C}
& \textbf{GPQA-D}
& \\

\midrule

0.0
& 45.5
& 68.5
& 24.7
& 46.2 \\

0.0005
& 47.0
& 69.5
& 27.8
& 48.1 \\

0.001
& \textbf{49.5}
& \textbf{70.5}
& 30.8
& \textbf{50.3} \\

0.005
& 46.5
& \textbf{70.5}
& \textbf{31.3}
& 49.4 \\

0.01
& 48.5
& \textbf{70.5}
& 27.3
& 48.8 \\

\bottomrule
\end{tabular}

\vspace{-0.1in}
\end{table}

\subsection{Computational Cost Analysis}
We compare the computational cost of \ours and TTRL using Qwen3-0.6B on the Math domain under the same experimental configuration on NVIDIA H200 GPUs. Each method is independently repeated three times, and we report the median runtime together with throughput, peak GPU memory, average GPU utilization, and energy consumption in \tabref{tab:runtime_cost}. Compared with TTRL, \ours incurs a 14.0\% runtime overhead and a 12.3\% reduction in throughput. This additional cost mainly arises from evaluating multiple parameter-perturbed policies during population exploration. \ours also shows 8.6 percentage points higher average GPU utilization and 24.0\% higher energy consumption. In contrast, peak GPU memory changes by less than 0.2\%. This is because \ours does not store all perturbed models in memory at once. Instead, each population member is represented by a random seed, and its perturbation is regenerated when needed through our memory-efficient perturbation implementation. As a result, only the anchor parameters and one temporary perturbation buffer are kept in memory at any time. Overall, \ours adds moderate computational overhead while keeping peak memory usage nearly unchanged.

\begin{table}[h]
\centering
\small
\caption{Runtime and computational cost comparison between \ours and TTRL using Qwen3-0.6B on the Math domain. Results are measured over three independent runs.}
\label{tab:runtime_cost}
\setlength{\tabcolsep}{5pt}
\renewcommand{\arraystretch}{1.05}

\resizebox{0.75\linewidth}{!}{%
\begin{tabular}{lccccc}
\toprule
\textbf{Method}
& \textbf{Runtime (min)}
& \textbf{Prompts/h}
& \textbf{Peak Mem. (GiB)}
& \textbf{GPU Util.}
& \textbf{Energy (Wh)} \\
\midrule
TTRL
& 78.85
& 304.38
& 100.94
& 67.6\%
& 303.3 \\

\ours
& 89.90
& 266.96
& 100.78
& 76.2\%
& 376.0 \\
\bottomrule
\end{tabular}%
}
\end{table}

\subsection{Robustness Across Random Seeds}

\tabref{tab:knowledge_variance} reports the performance of each method over three seeds. The results show that all methods exhibit relatively low variance and stable performance across runs.

\begin{table}[h]
\centering
\small
\caption{Pass@1 performance of different methods on the General Knowledge domain, reported as mean $\pm$ standard deviation over three seeds using Qwen3-0.6B.}
\label{tab:knowledge_variance}

\setlength{\tabcolsep}{6pt}
\renewcommand{\arraystretch}{1.1}

\begin{tabular}{lBGGc}
\toprule

\multirow{2}{*}{\textbf{Method}}
& \multicolumn{3}{c}{\textbf{General Knowledge}}
& \multirow{2}{*}{\textbf{Avg.}} \\

\cmidrule(lr){2-4}

& \textbf{MMLU}
& \textbf{ARC-C}
& \textbf{GPQA-D}
& \\

\midrule

Base model
& \(44.0_{\scriptscriptstyle \pm 0.0}\)
& \(65.5_{\scriptscriptstyle \pm 0.0}\)
& \(26.3_{\scriptscriptstyle \pm 0.3}\)
& \(45.3\) \\

TTRL
& \(44.5_{\scriptscriptstyle \pm 1.5}\)
& \(70.0_{\scriptscriptstyle \pm 1.6}\)
& \(25.8_{\scriptscriptstyle \pm 2.8}\)
& \(46.8\) \\

ECHO
& \(46.0_{\scriptscriptstyle \pm 1.0}\)
& \(69.5_{\scriptscriptstyle \pm 1.0}\)
& \(23.7_{\scriptscriptstyle \pm 1.1}\)
& \(46.4\) \\

Evol-RL
& \(46.5_{\scriptscriptstyle \pm 2.3}\)
& \(69.5_{\scriptscriptstyle \pm 0.3}\)
& \(27.3_{\scriptscriptstyle \pm 1.3}\)
& \(\underline{47.8}\) \\

ESRL
& \(49.5_{\scriptscriptstyle \pm 1.3}\)
& \(70.5_{\scriptscriptstyle \pm 0.3}\)
& \(30.8_{\scriptscriptstyle \pm 1.3}\)
& \(\mathbf{50.3}\) \\

\bottomrule
\end{tabular}

\vspace{-0.1in}
\end{table}

\newpage
\section{Failure Mode Discussion}

Hyperparameter settings play an important role in RL training, and their effects are further amplified in continual test-time learning, where supervision is derived entirely from self-generated signals. As shown in \tabref{tab:batch_size}, increasing the batch size to 128 results in only marginal gains over the base model. Unlike smaller batches, where sparse informative signals can have a strong influence on each update and drive rapid adaptation, an excessively large batch averages these signals with many weak or conflicting updates. This dilutes useful stochastic exploration and prevents newly discovered reasoning modes from being effectively reinforced. Consequently, excessively large batches substantially weaken the adaptation signal and lead to little improvement during continual test-time learning.

\begin{table}[h]
\vspace{-0.2in}
\centering
\small
\caption{Failed attempts with a large batch size using Qwen3-0.6B.}
\label{tab:batch_size}

\setlength{\tabcolsep}{6pt}
\renewcommand{\arraystretch}{1.05}

\begin{tabular}{cBGGc}
\toprule

\multirow{2}{*}{\textbf{Setting}}
& \multicolumn{3}{c}{\textbf{Math}}
& \multirow{2}{*}{\textbf{Avg.}} \\

\cmidrule(lr){2-4}

& \textbf{MATH-500}
& \textbf{AMC23}
& \textbf{AIME25}
& \\

\midrule

Base Model
& 5.0
& \textbf{6.0}
& \textbf{0.0}
& 3.7 \\

Batch Size = 128
& \textbf{6.0}
& \textbf{6.0}
& \textbf{0.0}
& \textbf{4.0} \\

\bottomrule
\end{tabular}

\end{table}

\begin{algorithm}[htp]
\caption{Evolution Strategy Reinforcement Learning (\ours)}
\label{alg:esrl}
\begin{algorithmic}[1]
\Require Unlabeled prompts $\mathcal{D}$; student parameters $\theta$;
frozen reference policy $\pi_{\theta_0}$; population size $N$;
perturbation radius $\sigma$; block size $B$; agreement threshold
$p_{\max}$; KL weight $\lambda$; learning rate $\eta$

\For{each block $\mathcal{B}$ of at most $B$ prompts in $\mathcal{D}$}
    \State $\theta_{\mathrm{anc}} \gets \theta$
    \Comment{Synchronise the anchor with the student}
    \State Sample perturbation seeds $s_1,\dots,s_N$

    \For{$k=1$ to $N$}
        \State $\varepsilon_k \gets \varepsilon(s_k)$
        \State $\theta_k \gets \theta_{\mathrm{anc}}
        + \sigma\varepsilon_k$
        \State $\{y_k^{(x)}\}_{x\in\mathcal{B}}
        \gets
        \textsc{GenerateBatch}
        (\pi_{\theta_k},\mathcal{B})$
    \EndFor

    \For{each prompt $x\in\mathcal{B}$}
        \State $\{a_k\}_{k=1}^{N}
        \gets
        \{\mathrm{Ans}(y_k^{(x)})\}_{k=1}^{N}$

        \State $a^\star,\{r_k\}_{k=1}^{N}
        \gets
        \textsc{MajorityVote}(\{a_k\}_{k=1}^{N})$
        \Comment{\eqref{eq:reward}}

        \If{\textsc{ValidityGate}$(\{a_k\}_{k=1}^{N},p_{\max})$
        is false}
            \State \textbf{continue}
        \EndIf

        \State $\{A_k\}_{k=1}^{N}
        \gets
        \textsc{Standardise}(\{r_k\}_{k=1}^{N})$
        \Comment{\eqref{eq:advantage}}

        \State $\theta
        \gets
        \theta-\eta\nabla_{\theta}\mathcal{L}(\theta)$
        \Comment{\eqref{eq:loss}}
    \EndFor
\EndFor

\end{algorithmic}
\end{algorithm}

\end{document}